\documentclass[sigconf]{acmart}

\usepackage{acmart-taps}
\usepackage{xcolor}

\copyrightyear{2026}
\acmYear{2026}
\setcopyright{cc}
\setcctype{by}
\acmConference[HUMA '26]{The 6th International Workshop on Human-centric Multimedia Analysis}{November 10--14, 2026}{Rio de Janeiro, Brazil}
\acmBooktitle{The 6th International Workshop on Human-centric Multimedia Analysis (HUMA '26), November 10--14, 2026, Rio de Janeiro, Brazil}
\acmDOI{10.1145/3841192.3841753}
\acmISBN{979-8-4007-2937-9/2026/11}

\begin{document}

\title[One Sensor, Whole Body]{One Sensor, Whole Body --- 3D Body Pose from a Single Consumer Earbud IMU}

\author{Zhilin Guo}
\authornote{These authors contributed equally to this work.}
\orcid{0000-0002-7660-3102}
\affiliation{%
  \institution{University of Cambridge}
  \country{United Kingdom}}

\author{Boqiao Zhang}
\authornotemark[1]
\orcid{0009-0000-0745-2438}
\affiliation{%
  \institution{University of Cambridge}
  \country{United Kingdom}}

\author{Oszkár Urbán}
\authornotemark[1]
\orcid{0009-0005-3552-0094}
\affiliation{%
  \institution{University of Cambridge}
  \country{United Kingdom}}

\author{Josef Bengtson}
\authornotemark[1]
\orcid{0000-0002-4321-491X}
\affiliation{%
  \institution{Chalmers University of Technology}
  \country{Sweden}}

\author{Hakan Aktas}
\orcid{0009-0009-4796-4281}
\affiliation{%
  \institution{University of Cambridge}
  \country{United Kingdom}}

\author{Wenzhao Li}
\orcid{0009-0007-0265-463X}
\affiliation{%
  \institution{University of Cambridge}
  \country{United Kingdom}}

\author{Siyu Hong}
\orcid{0009-0007-2347-9631}
\affiliation{%
  \institution{University of Cambridge}
  \country{United Kingdom}}

\author{Kyle Fogarty}
\orcid{0000-0002-1888-4006}
\affiliation{%
  \institution{University of Cambridge}
  \country{United Kingdom}}

\author{Chenliang Zhou}
\orcid{0009-0001-1096-1927}
\affiliation{%
  \institution{University of Cambridge}
  \country{United Kingdom}}

\author{Ali Senguel}
\orcid{0009-0000-9932-6200}
\affiliation{%
  \institution{University of Cambridge}
  \country{United Kingdom}}

\author{Cengiz Oztireli}
\correspondingauthor
\orcid{0000-0002-4700-2236}
\affiliation{%
  \institution{University of Cambridge}
  \country{United Kingdom}}
\email{aco41@cam.ac.uk}

\renewcommand{\shortauthors}{Guo et al.}

\begin{abstract}
Consumer earbuds already stream inertial motion data from the head, one of the most widely worn sensor locations on the body. We ask how much of the 3D body pose a single such head IMU can recover, and whether adding more consumer sensors actually helps. We build a multimodal capture pipeline that records four-view RGB-D video together with an AirPods head IMU and two Striv insole IMUs, synchronize the streams post-hoc, and generate pseudo-ground-truth with SAM 3D Body, yielding a 35-take single-subject benchmark spanning gait, turning, vertical, everyday, and clinically inspired motions. Adapting two recurrent model families (IMUPoser and MobilePoser), we show that one head IMU recovers lower-body pose at 79.0\,mm rigid-MPJPE and per-foot ground contact at 0.809 macro-F1, and that a causal variant retains most of this accuracy at streaming latency. In paired per-take significance tests across both families, adding the consumer foot IMUs never significantly improves pose and significantly degrades it in two of four model--split combinations; a mounting-bias probe and feet-only ablation identify insole orientation quality, not foot placement, as the mechanism. Extending the output to a 20-joint full-body skeleton maps the boundary: gross distal-arm motion is partially recoverable from the head alone, proximal upper-body pose is not, and staged fine-tuning recovers the leg accuracy that naive joint training sacrifices to multi-task dilution. For learned pose from consumer wearables, sensor reliability, not sensor count, is the binding constraint here. For the devices tested, the earbud is its sweet spot. Code is available at \url{https://github.com/ZhilinGuo/one-sensor-whole-body}.
\end{abstract}



\renewcommand{\shortauthors}{Guo et al.}

\maketitle

\section{Introduction}
Earbuds are among the most pervasive inertial sensor platforms: hundreds of millions already wear a calibrated IMU on the head for hours a day. If that single stream sufficed for useful 3D body pose, motion capture would need no suit, no straps, and no extra hardware. This paper asks two questions: \emph{how much of the} body pose can one consumer head IMU recover, and, because extra sensors are cheap to add but costly to wear, \emph{does adding more consumer sensors actually help}?

Most inertial pose systems assume far more instrumentation: six-IMU suits with a pelvis tracker~\cite{huang2018dip,yi2021transpose,yi2022pip,jiang2022tip}, VR headsets with 6-DoF SLAM positions and hand controllers~\cite{dittadi2021hmdfullbody,hmdposer,jiang2022avatarposer}, or phone--watch--earbud subsets evaluated on simulated data~\cite{imuposer,mobileposer}. Our setting is deliberately minimal and, to our knowledge, unoccupied on real data: a \emph{single, unified consumer earbud IMU}: 3-DoF fused orientation and acceleration, no absolute position, no hand trackers. Answering these questions requires labels a mocap lab cannot provide for everyday earbud wear, so we frame it as a human-centric multimodal capture problem. We record 35 takes spanning seven motion scripts covering both entertainment and clinically relevant movements, each with four-view RGB-D video plus head and foot IMU streams. With no hardware sync and no mocap lab, we align the modalities post-hoc using signal-based temporal calibration and use SAM 3D Body~\cite{sam3dbody} as pseudo-ground-truth (pseudo-GT), then evaluate IMUPoser-adapted~\cite{imuposer} and MobilePoser-adapted~\cite{mobileposer} recurrent models on a nine-joint lower-body skeleton, an auxiliary foot-contact task, a streaming (causal) variant, and a 20-joint full-body extension that maps the recoverability boundary. The study is a controlled pilot: one subject, one capture day, one earbud model, one insole product; all claims are within-participant and device-specific.

The answers are sharper than expected. (i) The head alone is strong: 79.0\,mm rigid-MPJPE on leave-one-run-out, 0.809 macro-F1 on per-foot contact, and a causal variant retaining most of this accuracy at one-frame latency. (ii) More sensors do not help: in paired per-take tests, adding two foot IMUs never significantly improves pose and significantly degrades it in two of four model--split combinations. (iii) The mechanism is reliability, not placement: a mounting-bias sweep and feet-only ablation trace the degradation to foot-orientation calibration, while the earbud's clean raw gyroscope remains directly learnable. (iv) The recoverability boundary is sharp: gross distal-arm swing carries signal, proximal upper-body pose does not, and staged fine-tuning recovers the leg accuracy that naive joint widening sacrifices. We contribute the capture pipeline and benchmark, paired-significance evidence that one head IMU is the optimal configuration among the consumer devices tested, a reliability probe explaining why, and a delimitation of what a single head IMU can and cannot recover.

\section{Related Work}
\noindent\textbf{Sparse-IMU full-body pose.}
Classic real-time methods (DIP, TransPose, PIP, and TIP) reconstruct full-body motion from six IMUs, one mounted at the pelvis or lower back~\cite{huang2018dip,yi2021transpose,yi2022pip,jiang2022tip}. Recent methods relax the count: IMUPoser uses variable subsets from phones, watches, and earbuds~\cite{imuposer}; MobilePoser targets 1--3 mobile IMUs~\cite{mobileposer}; others broaden layouts and generalization~\cite{zhang2024dynaip,vanwouwe2024diffusionposer}. Nearly all prior methods assume a pelvis sensor and raw gyroscopes, assumptions consumer earbuds and insoles cannot meet. Crucially, the commodity-device evaluations are synthetic (AMASS-simulated IMUs); we evaluate on real captured streams, where consumer noise and mounting variability, the phenomena at the heart of our finding, actually occur.

\noindent\textbf{Head-mounted minimal sensing.}
Dittadi et al.\ generate full-body SMPL poses from a single head-mounted device~\cite{dittadi2021hmdfullbody}; HMD-Poser scales from HMD-only to HMD plus a few IMUs~\cite{hmdposer}; AvatarPoser, AGRoL, EgoPoser, and DivaTrack hallucinate lower-body motion from head-and-hand trackers~\cite{jiang2022avatarposer,du2023agrol,jiang2024egoposer,yang2024divatrack}. These XR systems observe 6-DoF head \emph{position} from SLAM, usually with hand controllers, far richer than a raw earbud IMU. Ear2Pos reconstructs full-body pose from \emph{two} earbud IMUs with personalized bone lengths~\cite{ear2pos}, and ProgIP uses head-plus-wrists~\cite{progip}. We push minimalism to the limit: one unified head IMU stream (an earbud pair), no positions, no hands, no personalization, on real data.

\noindent\textbf{Foot/insole and clinical gait.}
Feet are established cues for gait phase and contact~\cite{sy2021liegroup,hori2025gip,grip2026,fog2025review}, but whether consumer insole IMUs stay reliable outside controlled straight-line locomotion is largely untested on real hardware.

\noindent\textbf{Vision+IMU and pseudo-GT.}
Where DeepFuse adds inertial signals to stabilize multi-view pose~\cite{huang2020deepfuse}, we use vision \emph{exclusively} to create labels: SAM 3D Body supplies pseudo-GT meshes~\cite{sam3dbody}, AMASS supplies synthetic pretraining IMUs~\cite{mahmood2019amass}, and temporal alignment follows camera--IMU calibration from motion agreement~\cite{li2014temporal}. No prior benchmark targets a single consumer earbud IMU with real multimodal capture and explicit sensor-reliability diagnostics.

\section{Capture System and Dataset}
\begin{figure}[t]
\centering
\includegraphics[width=0.40\linewidth]{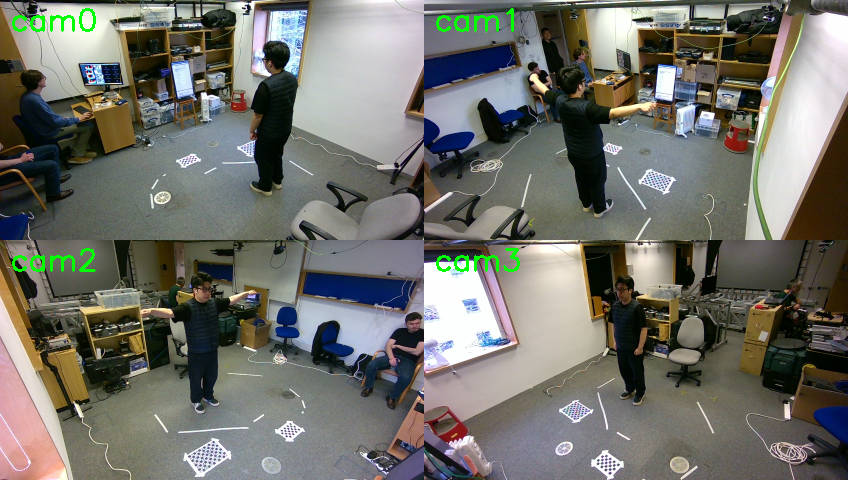}
\caption{One synchronized take from the four RealSense cameras; one view feeds SAM 3D Body pseudo-labeling, the others support qualitative checks and label fusion (Sec.~\ref{sec:results}).}
\Description{Four synchronized RGB camera views of a single motion take from the capture rig, showing the subject from different angles.}
\label{fig:cameras}
\end{figure}

\begin{table}[t]
\centering
\caption{Motion protocol: five runs repeat seven scripts (35 takes).}
\label{tab:protocol}
\scriptsize
\renewcommand{\arraystretch}{0.62}
\begin{tabular}{@{}c p{0.78\linewidth}@{}}
\toprule
Seq & Content \\
\midrule
S1 & Standing, T-pose, normal/slow/fast straight walking, $180^\circ$ turns \\
S2 & In-place stepping, marching, single-leg stance, weight shifts \\
S3 & Side-steps, $360^\circ$ in-place turns, backwards walking, figure-8 \\
S4 & Half/full squats, forward bends, lunges, sit-to-stand \\
S5 & Everyday movement: walk--sit--pick-up--reach \\
S6 & Clinical gait: turns, abrupt stop/restart, dual-task, sit-stand \\
S7 & Upper-limb reaches, pronation/supination, strikes, arm-swing \\
\bottomrule
\end{tabular}
\end{table}

The capture rig (Fig.~\ref{fig:cameras}) consists of four Intel RealSense D455e RGB-D cameras ($848\times480$ at 30\,FPS, color + depth), an AirPods earbud pair, exposed by iOS as a single fused head-motion stream, captured via a custom iOS/macOS app (fused attitude, device-frame acceleration, and gyroscope), and left/right Striv insole IMUs (fused Euler orientation plus acceleration, no raw gyroscope channel, ${\sim}30$\,Hz over BLE). The head IMU is our primary sensor; the insoles test whether extra consumer sensors earn their keep. The protocol (Table~\ref{tab:protocol}) comprises five repeated runs of seven motion scripts: 35 takes spanning gait, stepping, turning, vertical motion, everyday composites, clinically inspired gait, and upper-limb coordination, enabling leave-one-run-out and leave-one-motion-out evaluation.

\noindent\textbf{Dataset statistics.}

After processing and alignment, the benchmark contains 82{,}417 frames (45.8 minutes at 30\,FPS) across the 35 takes, averaging 78.5\,s each (38.6--137.1\,s). SAM 3D Body yields a usable lower-body label for every frame, and each foot is in ground contact for 65\% of frames (at least one foot for 73\%). Each take stores the aligned IMU streams, nine lower-body joints plus the full 70-keypoint SAM output, per-foot contact labels, and a validity mask.

\section{Alignment and Pseudo-Ground-Truth}
We do not utilise hardware synchronization, only Unix-style timestamps with constant offsets and slow drift. We place all streams on a common 30\,Hz grid via timestamps, refine the head offset by cross-correlating AirPods gyroscope magnitude against SAM-derived head angular speed, and refine a shared foot offset by matching Striv acceleration energy (heel-strike impacts) to SAM foot-speed minima. Across all 35 takes the recovered offsets are tightly clustered ($-2.69\pm0.06$\,s head, $-3.05\pm0.10$\,s feet), so misalignment is not silently absorbed into pose error. To generate pseudo ground-truth labels, we run SAM 3D Body on a selected camera view and convert its Momentum-Human-Rig output to a nine-joint lower-body target (pelvis, hips, knees, ankles, feet); the full 70-keypoint output is retained for the full-body extension (Sec.~\ref{sec:results}). We treat these as pseudo-GT, acknowledging single-view mesh recovery is less accurate than marker-based mocap.

\section{Benchmark Models and Protocol}

As an initial baseline, we adapt the two-layer bidirectional LSTM from IMUPoser~\cite{imuposer} to our nine-joint lower-body pose task. Each sensor contributes linear acceleration and a $3\times3$ orientation matrix (12-D per IMU; 36-D for head-plus-feet), with masked input blocks to support sensor configuration ablations. The model predicts 6D rotations for nine lower-body joints, whose positions are recovered via forward kinematics under SMPL-style conventions~\cite{loper2015smpl}. As a second baseline we adapt MobilePoser~\cite{mobileposer}, keeping its two-stage design and objectives (joint-position RNN feeding a pose RNN, teacher-forced noisy joints, smoothness/jerk penalties) retargeted to our sensors and lower-body output, and we attach an auxiliary foot-contact head to the IMUPoser-adapted backbone. We pretrain all models on synthetic IMU generated from CMU, BioMotionLab NTroje, and MPI HDM05 sequences in AMASS~\cite{mahmood2019amass}, then fine-tune on real pseudo-GT per fold. We evaluate under leave-one-run-out (held-out repeats of seen motion types) and leave-one-motion-out (held-out motion classes, the harder generalization test). Because per-frame Procrustes (PA-MPJPE) can favour a near-static mean pose, we emphasize \emph{rigid-MPJPE}, $\min_{s,R,\mathbf{t}}\frac{1}{TJ}\sum_{t,j}\| sR\,\hat{p}_{t,j}+\mathbf{t}-p_{t,j}\|_2$, which applies a \emph{single} similarity transform $(s,R,\mathbf{t})$ per sequence. We also report per-joint \emph{motion} $R^2_j = 1-\sum_t \|\hat{p}_{t,j}-p_{t,j}\|^2 / \sum_t \|p_{t,j}-\bar{p}_{j}\|^2$ against the temporal mean $\bar p_j$, so a static prediction scores ${\approx}\,0$. \emph{Distal} variants average knees, ankles, and feet, where foot sensors should matter most; PA-MPJPE and MPJVE are reported for completeness.

\noindent\textbf{Statistical protocol.}
Aggregate gaps between sensor configurations can hide fold noise, so every head-only vs.\ head+feet claim is backed by a \emph{paired per-take} analysis: rigid-MPJPE values are paired across the 35 takes and tested with Wilcoxon signed-rank (paired $t$-test secondary), Holm-corrected across the four model$\times$split comparisons, with effect sizes (median $\Delta$, Cohen's $d_z$) and win counts.

\noindent\textbf{Streaming and full-body variants.}
A \emph{causal} IMUPoser-adapted model (unidirectional LSTM, same capacity) emits each frame using only past observations, the configuration a deployed earbud system would run. A \emph{full-body} variant widens the output head to 20 SMPL joints, supervising the 17 with MHR70 pseudo-GT counterparts and the three spine joints weakly with interpolated targets (never evaluated); single-view arm labels are the noisiest source, so we down-weight upper-body loss, anchor the per-window alignment on the legs, and re-solve group alignment per body group in evaluation.

\noindent\textbf{Implementation.}
The IMUPoser-adapted model is a 512-unit two-layer BiLSTM (10.6M parameters); the MobilePoser-adapted model stacks two 256-unit two-layer BiLSTMs (5.3M total). Models train on 150-frame (5\,s) windows with Adam: synthetic pretraining 40 epochs (lr $3{\times}10^{-4}$), per-fold fine-tuning 20 epochs (lr $10^{-4}$). Because consumer calibration leaves a global frame/scale ambiguity, fine-tuning minimizes an L2 on forward-kinematics joints under a per-window closed-form rotation-plus-scale alignment (retraining takes a few hours on one A100).

\section{Results}
\label{sec:results}
\begin{table*}[t]
\centering
\caption{Full 35-take benchmark. Distal metrics average knees, ankles, and feet. Lower is better for errors, higher for $R^2$. ``run''/``seq'' = leave-one-run-out / leave-one-motion-out. Best per family in bold.}
{\scriptsize\renewcommand{\arraystretch}{0.58}\begin{tabular}{lcccccc}
\toprule
Method & rigid-MPJPE$\downarrow$ & $R^2_{\text{motion}}\uparrow$ & distal rigid$\downarrow$ & distal $R^2\uparrow$ & MPJVE$\downarrow$ & PA-MPJPE$\downarrow$ \\
\midrule
Constant mean-pose prior & 105.5 & 0.03 & 141.2 & -0.07 & - & - \\
Zero-input control (run) & 116.4 & -0.40 & 141.6 & -0.09 & 221 & 63.9 \\
\midrule
\multicolumn{7}{l}{\emph{IMUPoser-adapted}} \\
No calibration, head+feet (run) & 107.0 & -0.18 & 130.7 & 0.07 & 264 & 59.0 \\
\textbf{Head IMU only (run)} & \textbf{79.0} & 0.06 & 87.0 & 0.49 & 253 & 57.3 \\
Head IMU only (seq) & 92.6 & -0.10 & 107.2 & 0.27 & 249 & 64.3 \\
Head IMU only, causal (run) & 92.9 & -0.09 & 107.9 & 0.25 & 280 & 60.5 \\
Head IMU only, causal (seq) & 91.5 & -0.09 & 105.3 & 0.28 & 259 & 61.2 \\
Feet only (run) & 133.5 & -0.86 & 170.5 & -0.98 & 267 & 56.8 \\
Feet only (seq) & 129.7 & -0.69 & 164.2 & -0.68 & 292 & 66.8 \\
Head+feet (run) & 95.8 & -0.05 & 114.1 & 0.24 & 258 & 54.6 \\
Head+feet (seq) & 107.4 & -0.21 & 131.6 & -0.01 & 271 & 62.9 \\
\midrule
\multicolumn{7}{l}{\emph{MobilePoser-adapted}} \\
\textbf{Head IMU only (run)} & \textbf{85.5} & 0.05 & 98.7 & 0.37 & 258 & 58.9 \\
Head IMU only (seq) & 93.7 & -0.04 & 111.0 & 0.23 & 251 & 63.1 \\
Head+feet (run) & 86.9 & -0.02 & 97.9 & 0.43 & 277 & 59.2 \\
Head+feet (seq) & 99.1 & -0.21 & 116.3 & 0.15 & 265 & 67.1 \\
\bottomrule
\end{tabular}
}
\label{tab:main}
\end{table*}

\noindent\textbf{One head IMU is enough.}
Table~\ref{tab:main} summarizes the benchmark. Head-only is the best observed configuration in both families: 79.0\,mm rigid-MPJPE and 0.49 distal $R^2$ on leave-one-run-out for IMUPoser-adapted, generalizing better than every multi-sensor configuration when motion classes are held out (92.6 vs.\ 107.4\,mm; distal $R^2$ 0.27 vs.\ $-0.01$), and consistent across folds ($79.0\pm5.0$\,mm). A zero-input control (116.4\,mm) and mean-pose prior (105.5\,mm) confirm the model exploits real inertial signal. The result rests on both training stages (zero-shot transfer: 322.2\,mm; no pretraining: 94.1\,mm).

\noindent\textbf{Streaming readiness.}
Replacing the bidirectional LSTM with a unidirectional one (same capacity, one-frame latency) yields 92.9\,mm on leave-one-run-out and 91.5\,mm on leave-one-motion-out: the streaming head-only model still beats the \emph{bidirectional} head+feet configuration (95.8 and 107.4\,mm). With ${\sim}3{,}500$ CPU frames/s, a real-time earbud pose stream is deployable today.

\begin{figure}[t]
\centering
\includegraphics[width=0.66\linewidth]{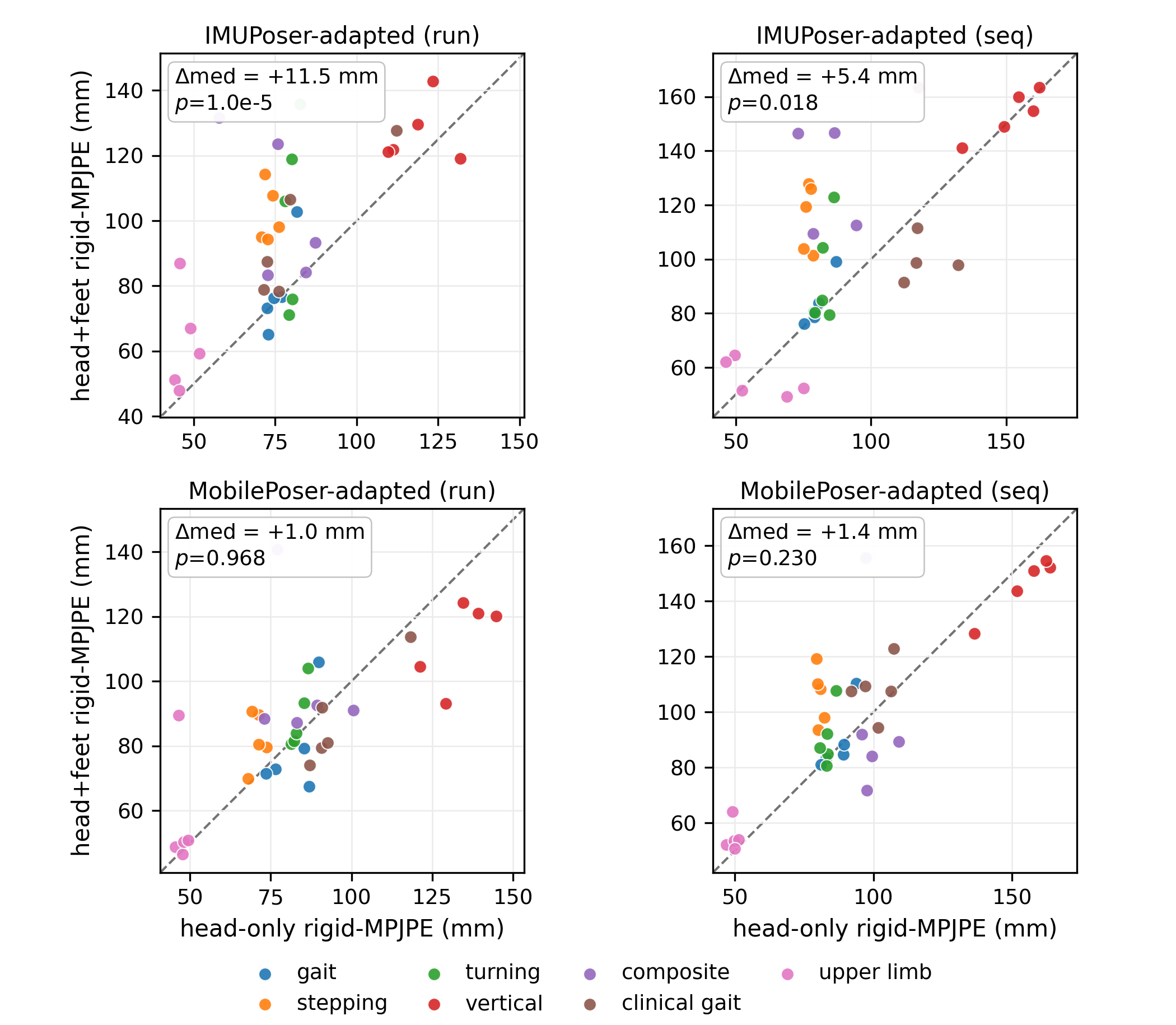}
\caption{Paired per-take head-only (x) vs.\ head+feet (y) rigid-MPJPE across both model families and splits; points above the diagonal favour head-only (systematic under IMUPoser-adapted, $p{<}0.02$ Holm; a tie-to-trend under MobilePoser-adapted).}
\Description{Four paired scatter plots comparing per-take rigid-MPJPE for head-only versus head-plus-feet sensor configurations, one panel per model family and evaluation split; points above the diagonal favour the head-only configuration.}
\label{fig:paired}
\end{figure}

\noindent\textbf{More sensors do not help: significantly.}
The paired per-take analysis (Fig.~\ref{fig:paired}) is the paper's central evidence. Across the 35 takes, head+feet never significantly beats head-only in any of the four model$\times$split comparisons. Under IMUPoser-adapted, head-only wins 29/35 takes on leave-one-run-out (median $\Delta$ $+11.5$\,mm, $d_z{=}0.88$, Wilcoxon $p{<}0.001$ Holm-corrected) and 24/35 on leave-one-motion-out ($+5.4$\,mm, $d_z{=}0.51$, $p{=}0.018$). Under MobilePoser-adapted the gap closes to a tie on held-out repeats ($p{=}0.97$) and a non-significant head-favouring trend on held-out motions ($p{=}0.23$). The feet-only configuration is the worst trained variant (133.5\,mm on held-out runs, worse than feeding no sensor at all), so the foot streams are individually weak \emph{and} damaging in combination. Discarding the foot IMUs is therefore a validated design decision, not discarded information.

\noindent\textbf{Why the feet fail: reliability, not placement.}
Three controls isolate the mechanism. (i) An explicit no-calibration variant with head+feet is worse still (107.0\,mm), implicating foot-orientation calibration. (ii) A mounting-bias probe injects a constant random sensor-to-bone rotation per take into the foot streams (per-axis std $\sigma$, three draws each). Realistic spread ($\sigma{=}5$--$10^\circ$) is near-neutral on average (92.6--92.8\,mm vs.\ 95.8\,mm uninjected) but highly variable across draws (85.0--98.7\,mm): individual sessions are luck-dependent. Beyond that, degradation is monotonic: 102.4$\pm$5.0\,mm at $\sigma{=}20^\circ$ and 118.9$\pm$1.2\,mm at $\sigma{=}40^\circ$, past even the 105.5\,mm mean-pose prior, because the network cannot down-weight a stream it no longer trusts. Even the luckiest draw remains worse than discarding the feet (79.0\,mm). (iii) The AirPods stream, by contrast, exposes a clean raw gyroscope, so head angular motion, correlated with stride, turning, and vertical excursions, is directly observed. The Striv insoles provide only a fused Euler orientation with no raw gyroscope, leaving a calibration residual the models cannot overcome. For this consumer device set, sensor \emph{reliability}, not sensor count, decides the outcome.

\noindent\textbf{Foot contact from the head alone.}
Foot-contact timing underpins gait segmentation, animation, and clinical timing measures. The auxiliary contact head, driven by the head IMU alone, reaches 0.809 macro-F1 over all 35 held-out takes while preserving pose accuracy (77.0\,mm rigid-MPJPE); it is easiest where feet are often stationary (upper-limb 0.968 F1) and hardest in turning (0.694). Temporal foot-state is recoverable from head motion alone even though foot orientation is too unreliable to help pose.

\noindent\textbf{The full-body boundary.}
Finally we widen the output to a 20-joint full-body skeleton (three seeds, leave-one-run-out). The boundary runs through the arm: the upper-body group beats its prior ($105.1{\pm}4.2$ vs.\ 131.5\,mm) with distal signal (elbows $R^2{=}0.45{\pm}0.06$, wrists $R^2{=}0.30{\pm}0.02$) tracking gross arm swing, while neck, head, and shoulders sit at or below the prior. This ceiling is not a label artifact: re-scoring held-out predictions against four-view-fused SAM 3D Body labels (per-frame lower-body-anchored similarity fusion, ten-take subset; cross-view residual 10--22\,mm) moves proximal-upper errors ${<}2$\,mm. Naive joint training, however, sacrifices the legs to the static-prior level ($107.5{\pm}13.1$\,mm vs.\ the 105.5\,mm prior; the dedicated model reaches 79.0\,mm). The cost is partly mechanical (unsupervised spine rotations corrupt the kinematic chain; weakly supervising them recovers 137.9\,$\to$\,115.1\,mm) and partly multi-task dilution from the arm labels. \emph{Staged fine-tuning} (ten lower-only epochs, then ten with the full weighted loss) removes most of the trade-off: the legs recover to $86.2{\pm}7.2$\,mm, within 7\,mm of the dedicated model, with the distal-arm signal intact (elbows $R^2{=}0.41$). Freezing the trunk after the lower stage protects the legs (85.7\,mm) but eliminates the arm signal ($R^2{\lesssim}0.05$; single-instance). On unseen motion classes the staged arm signal largely vanishes ($R^2{=}0.06$), and adding the feet hurts again (130.0\,mm legs, distal-arm $R^2$ ${\lesssim}0.2$), so the reliability thesis holds across output sets. One earbud IMU thus carries usable signal for lower-body pose \emph{and} gross arm swing via staged fine-tuning; a per-task model pair remains accuracy-optimal.

\section{Discussion and Limitations}
Our reading is constructive: head-only is not a compromise but the optimum of the consumer sensor set tested; the foot IMUs served their purpose by being falsified. Though foot sensors carry gait-phase and contact information, in the consumer setting this signal is unusable. We attribute the degradation to the quality and calibration of the \emph{consumer} insole orientation stream (Sec.~\ref{sec:results}). The pilot is limited to one subject, one capture day, and SAM 3D Body pseudo-GT rather than marker-based motion capture; our claims therefore establish within-participant feasibility for the tested hardware, not population-level generalization. A label-quality control against four-view-fused SAM 3D Body labels (Sec.~\ref{sec:results}) confirms single-view label noise does not drive the proximal ceiling, though the fused labels share the same backbone. Next steps: raw-gyroscope insole firmware and calibration, marker-based validation, multi-subject capture, and external baselines.

\section{Conclusion}
We asked how much of the 3D body pose a single consumer earbud IMU can recover, and built the capture pipeline and 35-take benchmark to answer it on real data. The answer: enough for lower-body pose (79.0\,mm rigid-MPJPE), foot contact (0.809 macro-F1), and, with staged fine-tuning, gross arm swing, at streaming latency; and never less, often significantly more, than with two extra foot IMUs. For the consumer devices tested, the design question is not sensor count, but which stream is reliable enough to learn from, and here that stream is the head.

\begin{acks}
This work was supported by a UKRI Future Leaders Fellowship [grant number G127262].
\end{acks}

\bibliographystyle{ACM-Reference-Format}
\bibliography{refs}

\end{document}